\documentclass[nohyperref]{article}

\usepackage{microtype}
\usepackage{graphicx}
\usepackage{subfigure}
\usepackage{booktabs} 
\usepackage{multirow}
\usepackage{algorithm}
\usepackage{algorithmic}

\usepackage{hyperref}

\usepackage[accepted]{icml2022_pods}

\usepackage{amsmath}
\usepackage{amssymb}
\usepackage{mathtools}
\usepackage{amsthm}

\usepackage[capitalize,noabbrev]{cleveref}

\theoremstyle{plain}

\theoremstyle{definition}

\theoremstyle{remark}

\usepackage[textsize=tiny]{todonotes}

\icmltitlerunning{Accepted to ICML Workshop on PODS 2022}

\begin{document}

\twocolumn[
\icmltitle{Estimating Test Performance for AI Medical Devices under Distribution Shift with Conformal Prediction}



\icmlsetsymbol{equal}{*}

\begin{icmlauthorlist}
\icmlauthor{Charles Lu}{equal,a}
\icmlauthor{Syed Rakin Ahmed}{equal,a,b}
\icmlauthor{Praveer Singh}{a,c}
\icmlauthor{Jayashree Kalpathy-Cramer}{a,c}
\end{icmlauthorlist}

\icmlaffiliation{a}{Department of Radiology, Massachusetts General Hospital}
\icmlaffiliation{b}{Department of Biophysics, Harvard University}
\icmlaffiliation{c}{Department of Ophthalmology, University of Colorado}

\icmlcorrespondingauthor{Charles Lu}{clu@mgh.harvard.edu}

\icmlkeywords{Machine Learning, ICML}

\vskip 0.3in
]



\printAffiliationsAndNotice{\icmlEqualContribution} 

\begin{abstract}
Estimating the test performance of software AI-based medical devices under distribution shifts is crucial for evaluating the safety, efficiency, and usability prior to clinical deployment.
Due to the nature of regulated medical device software and the difficulty in acquiring large amounts of labeled medical datasets, we consider the task of predicting the test accuracy of an arbitrary black-box model on an unlabeled target domain \textit{without} modification to the original training process or any distributional assumptions of the original source data (i.e. we treat the model as a ``black-box'' and only use the predicted output responses).
We propose a ``black-box'' test estimation technique based on conformal prediction and evaluate it against other methods on three medical imaging datasets (mammography, dermatology, and histopathology) under several clinically relevant types of distribution shift (institution, hardware scanner, atlas, hospital).
We hope that by promoting practical and effective estimation techniques for black-box models, manufacturers of medical devices will develop more standardized and realistic evaluation procedures to improve the robustness and trustworthiness of clinical AI tools.
\end{abstract}

\section{Introduction}\label{sec:intro}
Developing and evaluating AI healthcare software is extremely expensive due to the collection and curation of large amounts of medical data by clinical specialists. 
Even more costly is the process of submitting medical device software for regulatory approval, integrating it into existing clinical data infrastructure, training 
clinicians on appropriate usage, and continual post-deployment monitoring for failures.

Thus, it would be extremely valuable to be able to accurately estimate the prediction performance of models at new hospitals, patient populations, medical scanner equipment, etc. before actual clinical deployment.
From a regulatory point of view, unusually low (or high) test performance could indicate unidentified or unmitigated risks that could degrade the quality of AI medical devices in different scenarios.
For example, drift in disease prevalence over time might be quantified and necessitate periodic re-calibration of the AI algorithm by the device manufacturer.

This problem of trying to identify and rectify performance degradation under new data populations has been extensively studied as distribution shift, out-of-distribution detection, and domain generalization~\cite{DBLP:journals/corr/abs-2110-11334,DBLP:journals/corr/abs-2103-02503}.
Recent works have begun to investigate techniques and frameworks for estimating test performance on unlabeled, domain-shifted distribution.

\citet{DBLP:journals/corr/abs-2007-02915} introduced the notion of predicting performance on an unlabeled test set and, for this task, evaluated a regression approach using feature vectors from models trained under different distribution shifts. 
A simpler and more practical technique for estimating accuracy on an unlabeled target distribution was proposed by \citet{DBLP:journals/corr/abs-2201-04234}, which selects a confidence threshold using accuracy on a source dataset.

In this paper, we propose a novel estimation technique based on conformal prediction and evaluate it against other ``black-box'' test estimation methods on several medical imaging datasets.

\section{Estimating ``Black-box'' Test Performance}\label{sec:methods}
We make the assumption that the model of a software medical device will be ``static'' and not directly accessible.
Therefore, for an estimation test method to be valid for an arbitrary black-box model, we assume only access to predicted outputs from the model from which to estimate performance.
For example, for a multi-class classification model, $F$, we might only receive a predicted response of probability scores, $\Delta^{\lvert \mathcal{Y} \rvert}$, where ${\lvert \mathcal{Y} \rvert}$ is the number of classes and $\Delta$ is the probability simplex. Additionally, we assume we can access to a portion of the source distribution's labeled test set, $\{(x, y)\}_{i=1}^m \sim \mathcal{D}^S$, to use to estimate the target distribution \textit{unlabeled} test set $\{x\}_{i=1}^n \sim \mathcal{D}^T$.

While there are existing techniques for domain adaption and distribution shift, they necessarily assume a modifiable training procedure or preexisting knowledge of the type or character of distribution shift (e.g. label shift, covariate shift, etc.)~\cite{DBLP:journals/corr/abs-1802-03916}.
Absent these restrictions, our objective is to empirically evaluate arbitrary black-box models for medical applications that might be easily implemented into current deployment practices for AI medical devices or otherwise useful in informing regulatory testing and monitoring of machine learning algorithms.

A first intuitive approach to estimating test performance on target data would be to use the maximum softmax probability as an estimate of accuracy~\cite{DBLP:journals/corr/HendrycksG16c}.
While deep learning models have been found to be poorly calibrated~\cite{DBLP:journals/corr/GuoPSW17}, more recent work finds newer architectures to be better calibrated than previously thought~\cite{DBLP:journals/corr/abs-2106-07998}.
Therefore, one method to estimate test accuracy is to calculate the \textit{average confidence} (AC) score on the unlabeled target test dataset: 
\begin{equation}
    \small
    \text{AC}(X^T) = \mathbb{E}_{x\sim\mathcal{D}^T}\left[{\max}_{j \in \mathcal{Y}} F(x)_j\right],
\end{equation}
where $j$ is the index of a particular label's softmax score.

Another general approach involves computing the difference between source and target distributions.
\citet{DBLP:journals/corr/abs-2107-03315} 
proposes one method, which does not require model access, that estimates target accuracy with the source accuracy plus the \textit{difference in confidence} (DOC) between source and target datasets:
\begin{equation}
    \small
    \begin{split}
        DOC(X^S, Y^S, X^T) = \mathbb{E}_{(x, y)\sim\mathcal{D}^S}\left[\mathbb{I}\left[{\arg\max}_{y\in\mathcal{Y}} F(x)_j = y \right]\right] \\ + \left|\mathbb{E}_{x\sim\mathcal{D}^T}\left[\max_{j \in \mathcal{Y}} F(x)_j\right] - \mathbb{E}_{x\sim\mathcal{D}^S}\left[\max_{j \in \mathcal{Y}} F(x)_j\right]\right|, 
    \end{split}
\end{equation}
where $\mathbb{I}$ is the indicator function.

A third approach is to choose a score threshold on the source distribution to estimate the test accuracy on the target distribution. 
\textit{Average threshold confidence} (ATC) is a technique that estimates test accuracy as the expected number of target data points above a threshold: 
\begin{equation}
    \small
    ATC(X^T) = \mathbb{E}_{x \sim\mathcal{D}^T}\left[\mathbb{I}\left[\max_{j \in \mathcal{Y}} s(x)_j > t\right]\right],
\end{equation} 
where $t$ is a threshold estimated on some score function, $s: \Delta^k \rightarrow \mathbb{R}$ function.
\citet{DBLP:journals/corr/abs-2201-04234} proposes to choose $t$ such that the percentage of data points with a score above the threshold is equal to the accuracy on the source dataset: $\mathbb{E}_{x\sim\mathcal{D}^S}\left[\mathbb{I}\left[s(x) > t \right]\right] = \mathbb{E}_{(x,y)\sim\mathcal{D}^S}\left[\mathbb{I}\left[{\arg\max}_{j\in\mathcal{Y}} F(x)_j = y\right]\right]$.

\subsection{Estimating Test Accuracy with Conformal Prediction}\label{sec:estimate}
Conformal prediction is an approach to distribution-free uncertainty quantification that provides marginal coverage guarantees on arbitrary models, such as deep convolutional neural networks~\cite{DBLP:journals/corr/abs-2107-07511}.
In conformal prediction, a score threshold, $\hat{t}$, is estimated such that the prediction set, $\mathcal{T}(x) = \{j \in \mathcal{Y} \mid s(x)_j > \hat{t}\}$, contains the true label on average,
$P\left(Y \in \mathcal{T}(X)\right) \geq 1 - \alpha$,
where $\alpha$ is some user-specified confidence level.
\cite{10.5555/1062391}.
Therefore a prediction on a specific data point will be a set of plausible classes instead of only that class with the highest softmax score.
The marginal coverage guarantee of conformal prediction assumes that $t$ is calibrated on a held-out set drawn exchangeably from the test distribution.
Naturally, this assumption is violated under distribution shift. 
While conformal prediction can be modified if the type of shift is known~\cite{https://doi.org/10.48550/arxiv.1904.06019,podkopaev2021distributionfree}, we find that base conformal prediction by itself to be a useful measure of uncertainty by which to estimate test accuracy on the target distribution.
The basic intuition is that the more data points with large prediction set sizes, the more difficult for the model to classify these points, and subsequently produce a lower estimated test performance. 

We propose conformal prediction confidence (CPC) to estimate test accuracy on unlabeled target datasets.
We use the conformal prediction method by~\citet{doi:10.1080/01621459.2017.1395341} and estimate $\hat{t}$ using a small subset of source data and evaluate two methods of conformal prediction confidence by setting $\alpha$ to be the source accuracy (CPC-ACC) or the average confidence of the target dataset (CPC-AC).
To estimate test accuracy on target distribution, we take the expected value of the average softmax score of each prediction set in the target dataset to be the estimated test accuracy.
\begin{equation}
    \small
    CPC(X^T) = \mathbb{E}_{x \sim\mathcal{D}^T}\left[ \mathbb{E}\left[\{s(x)_j \mid j \in \mathcal{T}(x) \}\right]\right]
\end{equation} 
See Algorithm~\ref{alg:cp} for implementation details of CPC-ACC.

Intuitively, more confident prediction will have smaller set sizes and vice versa.
A highly confident prediction may have only one element in its set with a high score while a less confident prediction may have many elements in the prediction set.
Thus, we would expect the average softmax score for low confidence predictions (sets with many items) to be less than high confidence predictions (sets with few items with high scores).
Then the average of these low and high confidence scores can be taken as an estimate of overall accuracy (in the special case of every prediction set only having one item, namely the maximum score, CPC-AC is equivalent to AC).

\section{Experiments}\label{sec:experiments}

\begin{figure*}[h!]
\centering
\includegraphics[width=0.92\textwidth]{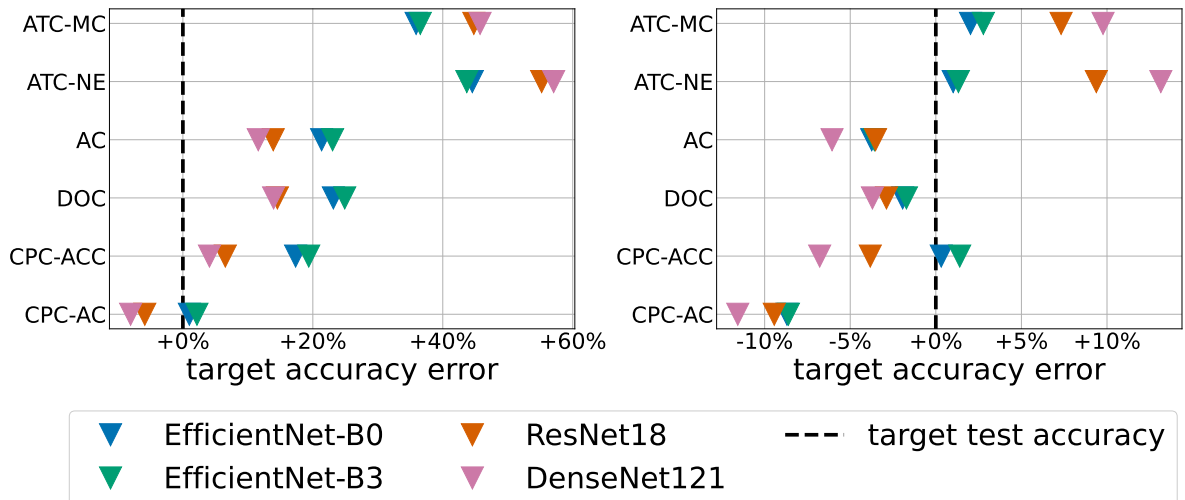}
\caption{Comparison of test estimation methods using four different architectures on Fitzpatrick dataset (atlas 1 $\rightarrow$ atlas 2); left if without temperature scaling; right is with temperature scaling.}
\label{fig:fitzpatrick}
\end{figure*}

\begin{table*}[h!]
\caption{Comparison by dataset and distribution shift using EfficientNet-B0 averaged over 5 training runs.} \label{tab:results}
\begin{small}
\begin{sc}
\begin{tabular}{lllll}
\toprule
dataset & source $\rightarrow$ target (accuracy \%) & \multirow{2}{*}{method}  &  \multicolumn{2}{c}{absolute target error} \\
\cline{4-5}
& & & no scaling & with scaling \\
\midrule
\multirow{6}{*}{CIFAR-10} 
    & \multirow{6}{*}{original $(83\%)$ $\rightarrow$ CIFAR-10.1 $(71\%)$}
        &   ATC-MC         &  $ 19.8\% \pm 1.4\% $ & $ 1.6\% \pm 0.7\% $ \\
        & & ATC-NE         &  $ 25.3\% \pm 1.2\% $ & $ \bf{0.7\% \pm 0.7\%} $ \\
        & &  AC            &  $ 19.7\% \pm 0.9\% $ & $ 6.0\% \pm 0.5\% $ \\
        & & DOC            &  $ 18.4\% \pm 1.2\% $ & $ 4.8\% \pm 0.7\% $ \\
        & & CPC-ACC  &  $ 19.9\% \pm 0.8\% $ & $ 8.6\% \pm 0.6\% $ \\
        & & CPC-AC   &  $\bf{16.2\% \pm 1.0\%}$ & $15.7\% \pm 1.5\% $ \\
\midrule
\multirow{6}{*}{Fitzpatrick17K} 
    & \multirow{6}{*}{atlas 1 $(37\%)$ $\rightarrow$ atlas 2 $(51\%)$}
        &   ATC-MC         &  $ 35.9\% \pm 0.7\% $ & $ 2.2\% \pm 2.3\% $ \\
        & & ATC-NE         &  $ 44.5\% \pm 0.6\% $ & $ 2.7\% \pm 2.3\% $ \\
        & &  AC            &  $ 21.4\% \pm 1.0\% $ & $ 3.7\% \pm 1.7\% $ \\
        & & DOC            &  $ 23.2\% \pm 0.8\% $ & $ 2.4\% \pm 1.1\% $ \\
        & & CPC-ACC  &  $ 17.3\% \pm 1.1\% $ & $ \bf{0.8\% \pm 0.5\%}$ \\
        & & CPC-AC   &  $ \bf{1.1\% \pm 0.7\%}$ & $ 8.7\% \pm 1.0\% $ \\
\midrule
\multirow{24}{*}{DMIST} 
    & \multirow{6}{*}{scan. 1 $(70\%$) $\rightarrow$ scan. 2 $(58\%)$}
        &   ATC-MC         &  $     26.8\% \pm 6.5\% $ & $ \bf{4.8\% \pm 2.0\%} $ \\
        & & ATC-NE         &  $     33.5\% \pm 9.6\% $ & $     4.9\% \pm 1.7\%  $ \\
        & &  AC            &  $     25.7\% \pm 3.6\% $ & $    11.8\% \pm12.1\%  $ \\
        & & DOC            &  $     22.7\% \pm 8.9\% $ & $     8.8\% \pm 0.5\%  $ \\
        & & CPC-ACC  &  $     26.0\% \pm 3.8\% $ & $    13.2\% \pm11.6\%  $ \\
        & & CPC-AC   &  $ \bf{21.4\% \pm 3.3\%}$ & $     9.3\% \pm 0.6\%  $ \\
        \cmidrule{2-5}
    & \multirow{6}{*}{scan. 1 $(70\%)$ $\rightarrow$ scan. 3 $(68\%)$}
        &   ATC-MC         &  $     22.5\% \pm 1.8\% $ & $     3.2\% \pm 1.6\%  $ \\
        & & ATC-NE         &  $     28.7\% \pm 4.4\% $ & $     3.9\% \pm 1.3\%  $ \\
        & &  AC            &  $     19.2\% \pm 3.6\% $ & $     6.7\% \pm11.3\%  $ \\
        & & DOC            &  $ \bf{16.7\% \pm 7.8\%}$ & $ \bf{2.4\% \pm 1.0\%} $ \\
        & & CPC-ACC  &  $     19.4\% \pm 3.9\% $ & $     6.9\% \pm11.5\%  $ \\
        & & CPC-AC   &  $     17.1\% \pm 2.2\% $ & $     3.8\% \pm 5.1\%  $ \\
        \cmidrule{2-5}
    & \multirow{6}{*}{scan. 1 $(70\%)$ $\rightarrow$ scan. 4 $(51\%)$}
        &   ATC-MC         &  $     32.1\% \pm 7.2\% $ & $     5.6\% \pm 2.6\%  $ \\
        & & ATC-NE         &  $     37.4\% \pm 9.5\% $ & $ \bf{3.6\% \pm 4.0\%} $ \\
        & &  AC            &  $     31.8\% \pm 4.1\% $ & $    17.3\% \pm12.5\%  $ \\
        & & DOC            &  $     28.8\% \pm 9.8\% $ & $    14.3\% \pm 2.2\%  $ \\
        & & CPC-ACC  &  $     32.7\% \pm 4.3\% $ & $    20.1\% \pm11.6\%  $ \\
        & & CPC-AC   &  $ \bf{24.4\% \pm11.0\%}$ & $    14.4\% \pm 7.5\%  $ \\
        \cmidrule{2-5}
    & \multirow{6}{*}{DMIST $(67\%)$ $\rightarrow$ external inst. $(52\%)$}
        &   ATC-MC         &  $     35.8\% \pm 5.6\% $ & $     9.6\% \pm 5.3\%  $ \\
        & & ATC-NE         &  $     38.8\% \pm 6.2\% $ & $    15.8\% \pm 5.6\%  $ \\
        & &  AC            &  $     36.8\% \pm 2.0\% $ & $    15.7\% \pm12.9\%  $ \\
        & & DOC            &  $ \bf{20.5\% \pm15.6\%}$ & $ \bf{8.2\% \pm 4.2\%} $ \\
        & & CPC-ACC  &  $     26.3\% \pm 2.7\% $ & $     16.8\% \pm11.2\% $ \\
        & & CPC-AC   &  $     21.1\% \pm 7.3\% $ & $     11.7\% \pm 4.1\% $ \\
\midrule
\multirow{6}{*}{Camelyon17} 
    & \multirow{6}{*}{hospital 1 $(99\%)$ $\rightarrow$ hospital 2 $(85\%)$}
        &   ATC-MC         &  $    12.9\% \pm 3.0\% $ & $     8.0\% \pm 5.3\% $ \\
        & & ATC-NE         &  $    12.9\% \pm 3.0\% $ & $     8.0\% \pm 5.6\% $ \\
        & &  AC            &  $     9.8\% \pm 2.8\% $ & $     7.7\% \pm12.9\% $ \\
        & & DOC            &  $    20.9\% \pm 7.5\% $ & $ \bf{4.1\% \pm 1.6\%}$ \\
        & & CPC-ACC  &  $ \bf{8.8\% \pm 2.8\%}$ & $     7.9\% \pm 4.2\% $ \\
        & & CPC-AC   &  $    11.8\% \pm 1.3\% $ & $     4.2\% \pm 3.5\% $ \\
\bottomrule
\end{tabular}
\end{sc}
\end{small}
\vskip -0.1in
\end{table*}

We conduct experiments with four datasets with the associated distribution shifts:
\begin{itemize}
    \item \textbf{CIFAR-10}~\cite{Krizhevsky09learningmultiple} -- The source dataset is the original CIFAR-10 (25,000 train / 25,000 validation / 10,000 test), while the target dataset is CIFAR-10.1~\cite{recht2018cifar10.1} (2,000 test, 10 classes of 62,000 natural images)
    \item \textbf{Fitzpatrick17K}~\cite{groh2021evaluating} -- The source dataset is the DermaAmin atlas (7295 train / 811 validation / 3474 test), while the target dataset is Atlas Dermatologico (3889 test, 114 lesion classes in dermatology images)
    \item \textbf{DMIST}~\cite{doi:10.1056/NEJMoa052911} -- Distribution shift is evaluated between four scanner types (1: 12421, 2: 47896, 3: 41311, 4: 6562) and from DMIST (33 institutions) to an external healthcare institution (10,819 test) in a multi-class breast density rating task of mammograms.
    \item \textbf{WILDS-Camelyon17}~\cite{DBLP:journals/corr/abs-2112-05090,bandi2018detection} -- Distribution shift is evaluated under source hospitals (302,436 train / 34,904 validation / 33,560 test) to target hospital (85,054 test) for metastases detection (binary) from histopathology.
\end{itemize}

We evaluate a variety of network architectures and perform five training runs for each model.
Additionally, we compare results with and without temperature scaling (TS) to study the effects of softmax calibration on test estimation methods (although access to the original logits will not likely be available in most commercial models)~\cite{DBLP:journals/corr/GuoPSW17}.

We report state-of-the-art test accuracy prediction using CPC on several medical imaging datasets without TS and competitive performance post-TS, summarized in Table~\ref{tab:results}.
The performance gains with CPC are especially noticeable on the Fitzpatrick17K dataset, which contains a much larger number of classes compared to other datasets.

Interestingly, we observe that TS moderates the effect of overestimation of predicted test accuracy across methods and architectures, as can be seen in Figures~\ref{fig:fitzpatrick}--~\ref{fig:dmist}.
We hypothesize that further analysis of the effect of model architecture on different medical datasets and tasks will be crucial in developing robust and reliable performance estimation techniques for medical applications.
Other follow-up work might extend test estimation methods to other clinically relevant tasks such as semantic segmentation, medical image registration, and automatic medical report classification.

\section{Clinical Relevance}\label{sec:clinic}
When considering clinical deployment, the majority of work incorporating AI for clinical classification tasks has traditionally focused on model accuracy and classification performance on in-distribution (ID) data; these models tend to break down when inferring on out-of-distribution (OOD) data, which could be images acquired from different cameras, scanners, and/or institutions that may have different inherent modality, view, and/or quality. This brittleness of models is evidenced by examples of one-pixel attacks~\cite{su2019one} and pose-perturbed images~\cite{alcorn2019strike} fooling neural networks. In the clinical realm, this imposes considerable concern and distress: ``black-box'' models that misclassify OOD data can incorrectly disqualify patients from clinical trials or result in unnecessary procedures. This is particularly relevant for models designed for radiogenomic prediction in tumors, where incorrect classification of a tumor-specific genetic marker status could, at worst, prevent patients from accessing the appropriate targeted therapy.

A second concern is the lack of definitive ground truth and uncertainty with labeling data in specific clinical domains. Prominent examples of this include the significant inter-rater variability in ascertaining the pre-neoplastic or neoplastic status of cervical lesions and the possible presence of intra/inter-lesional genetic heterogeneity for brain tumors, both of which can cause ``black-box'' AI models to fail on external data. Approaches that create site-specific or dataset-specific models are often unfeasible for these clinical tasks since this adds to the complexity of having to train and deploy multiple models for the same task; multiple networks also pose additional logistical concerns when seeking FDA approval for deployment.

\bibliography{pods}
\bibliographystyle{icml2022}

\newpage
\appendix
\onecolumn
\section{Appendix}

\begin{figure*}[h!]
\centering
\includegraphics[width=0.92\textwidth]{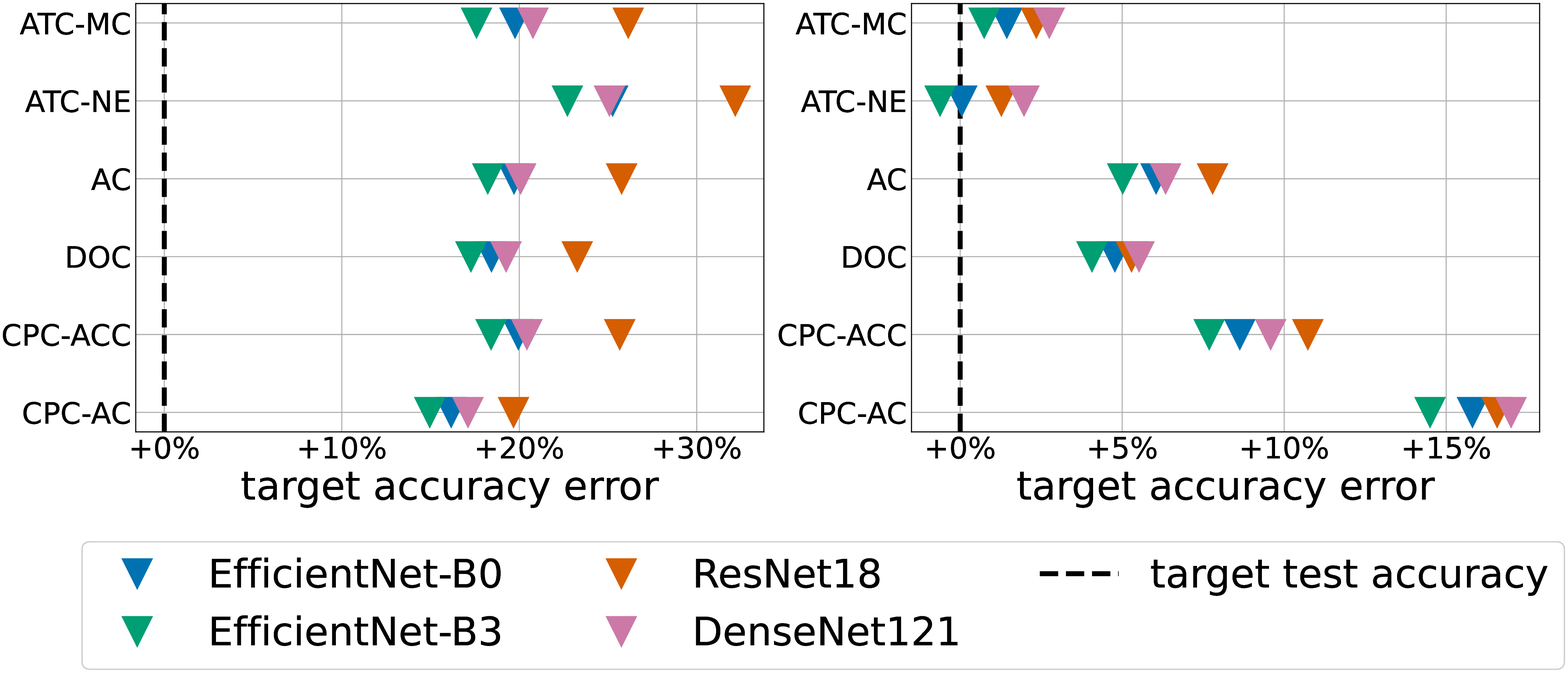}
\caption{Comparison of test estimation methods using four different architectures on CIFAR-10 dataset (original test set $\rightarrow$ CIFAR-10-1 test set); left if without temperature scaling; right is with temperature scaling.}
\label{fig:cifar}
\end{figure*}

\begin{figure*}[h!]
\centering
\includegraphics[width=0.92\textwidth]{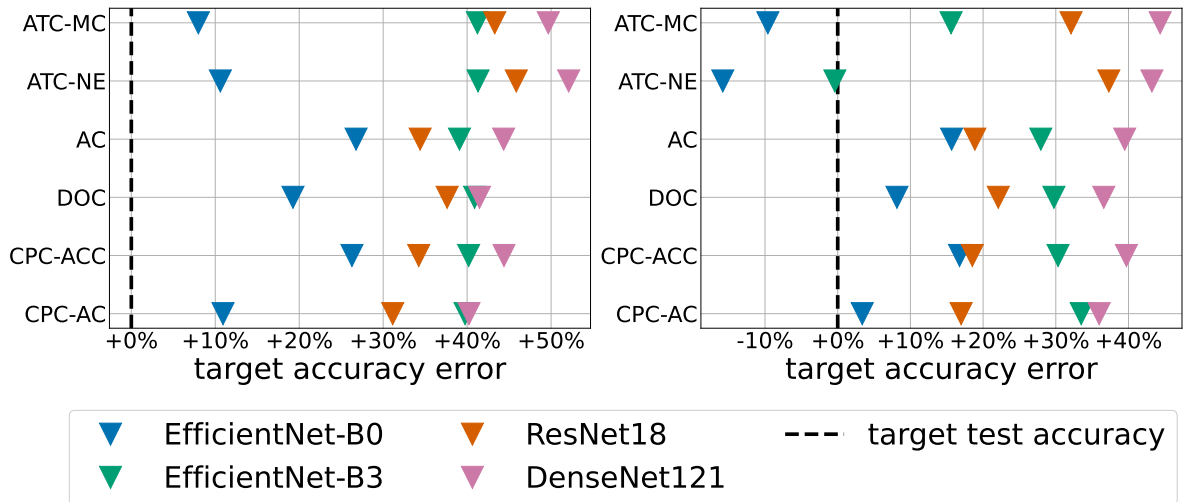}
\caption{Comparison of test estimation methods using four different architectures on DMIST dataset (internal institutions $\rightarrow$ external institution); left if without temperature scaling; right is with temperature scaling.}
\label{fig:dmist}
\end{figure*}

\begin{figure*}[h!]
\centering
\includegraphics[width=0.92\textwidth]{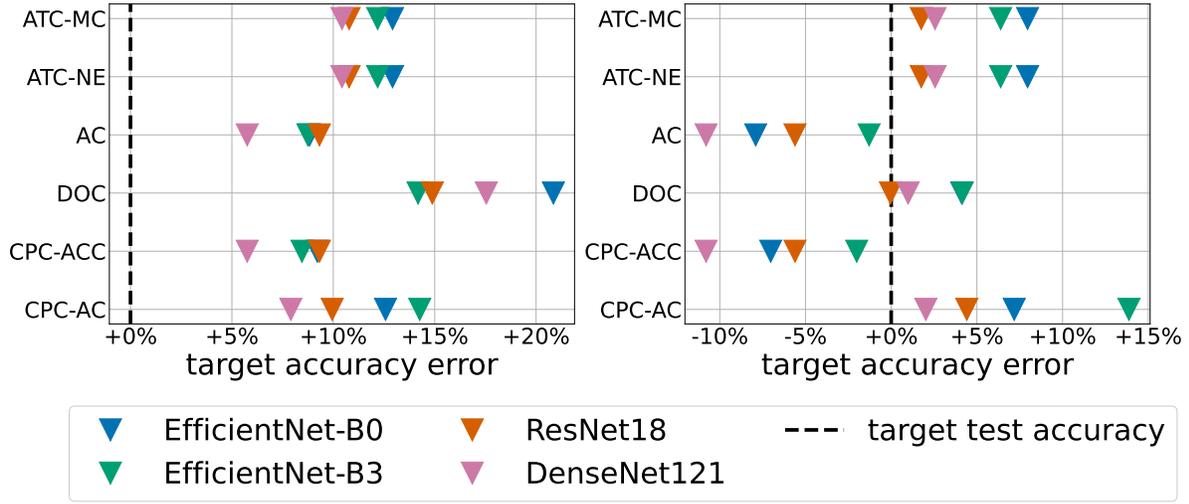}
\caption{Comparison of test estimation methods using four different architectures on Camelyon dataset (hospital 1 $\rightarrow$ hospital 2); left if without temperature scaling; right is with temperature scaling.}
\label{fig:camelyon}
\end{figure*}

\begin{algorithm}[h!]
    \caption{Conformal prediction confidence using source accuracy (CPC-ACC)}
    \label{alg:cp}
    \textbf{Input:} \\
    source calibration set $\{(x_i, y_i)\}_{i=1}^k \sim \mathcal{D}^S$, \\
    target test set $\{(x_i, y_i)\}_{i=1}^n \sim \mathcal{D}^T$, \\
    confidence level $\alpha = \mathbb{E}_{(x,y)\sim\mathcal{D}^S}\left[\mathbb{I}\left[{\arg\max}_{j\in\mathcal{Y}} f(x)_j = y\right]\right]$, \\
    model $F: X \rightarrow \Delta^{\lvert\mathcal{Y}\rvert}$, \\
    quantile function $Q: \{\mathbb{R}\}^m \times [0, 1] \rightarrow \mathbb{R}$
    \begin{algorithmic}[1] 
        \STATE $c \leftarrow \{\}$
        \FOR{$i \in \{1, 2, \ldots m\}$}
            \STATE $s_\text{cal} \leftarrow s_\text{cal} \cap \max F(x_i)$
        \ENDFOR
        \STATE $\hat{t} \leftarrow Q(s, \frac{\lceil \alpha \cdot (m + 1) \rceil}{m})$
        \STATE $p \leftarrow 0$
        \FOR{$i \in \{1, 2, \ldots n\}$}
            \STATE $s \leftarrow F(X_i)$
            \STATE $\tau \leftarrow \{j \in \mathcal{Y} \mid s_j > \hat{t}\}$
            \STATE $p \leftarrow p + \frac{1}{\lvert\tau\rvert} \sum_{j \in \tau} s_j $
        \ENDFOR
        \STATE $p \leftarrow p / n$
    \end{algorithmic}
    \textbf{Output}: estimated target accuracy $p$ \\
\end{algorithm}


\end{document}